TUM School of Computation, Information and Technology
Technical University of Munich

Bachelor's Thesis in Computer Science

# Evaluation of Differential Privacy Mechanisms on Federated Learning

Evaluierung von Differential Privacy Mechanismen im Föderierten Lernen

**Supervisor**     Prof. Dr.-Ing. habil. Alois C. Knoll

**Advisor**        Nagacharan Teja Tangirala, M.Sc.

**Author**         Tejash Varsani

**Date**           March 27, 2025 in Munich

# Disclaimer

I confirm that this Bachelor's Thesis is my own work and I have documented all sources and material used.

Munich, March 27, 2025
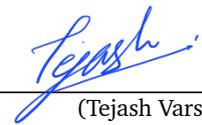
(Tejash Varsani)

## Abstract

Federated learning is distributed model training across several clients without disclosing raw data. Despite the advancements in data privacy, risks still remain. Differential Privacy (DP) is a technique to privatize sensitive data. A common DP approach is adding noise to model updates, usually controlled by a fixed privacy budget. However, this approach can produce too much noise, particularly when the model converges, compromising performance. To address this problem, the adaptive privacy budget has recently been investigated as a solution. This work implements DP methods using Laplace and Gaussian mechanisms with an adaptive privacy budget, extending the $SelecEval$ simulator. We introduce an adaptive clipping approach in the Gaussian mechanisms, ensuring that gradients of the model are dynamically updated rather than using a fixed sensitivity. We conduct extensive experiments with various privacy budgets, $IID$ and $non-IID$ datasets, and different numbers of selected clients per round. While our experiments were limited to 200 training rounds, our results suggest that adaptive privacy budgets and adaptive clipping can help maintain model accuracy while preserving privacy.

## Kurzfassung

Federated Learning ist das verteilte Trainieren von Modellen über mehrere Clients hinweg, ohne dass Rohdaten offengelegt werden. Trotz Fortschritten im Bereich Datenschutz bestehen weiterhin Risiken. Differential Privacy (DP) ist eine Technik zur Privatisierung sensibler Daten. Ein gängiger DP-Ansatz besteht darin, Rauschen zu den Modell-Updates hinzuzufügen, was üblicherweise durch ein festgelegtes Privacy-Budget gesteuert wird. Dieser Ansatz kann jedoch insbesondere bei der Konvergenz des Modells zu viel Rauschen erzeugen und dadurch die Leistung beeinträchtigen. Um dieses Problem zu lösen, wurde kürzlich das adaptive Privacy-Budget als mögliche Lösung untersucht. In dieser Arbeit werden DP-Methoden unter Verwendung von Laplace- und Gaußschen Mechanismen mit adaptivem Privacy-Budget implementiert, wobei der $SelecEval$-Simulator erweitert wurde. Wir führen zudem einen adaptiven Clipping-Ansatz in die Gaußschen Mechanismen ein, der sicherstellt, dass die Gradienten des Modells dynamisch angepasst werden, anstatt eine feste Sensitivität zu verwenden. Wir führen umfassende Experimente mit verschiedenen Privacy-Budgets, IID- und nicht-IID-Datensätzen sowie unterschiedlichen Anzahlen an ausgewählten Clients pro Runde durch. Obwohl unsere Experimente auf 200 Trainingsrunden beschränkt waren, deuten die Ergebnisse darauf hin, dass adaptive Privacy-Budgets und adaptives Clipping dazu beitragen können, die Modellgenauigkeit aufrechtzuerhalten und gleichzeitig die Privatsphäre zu schützen.

# Contents

# Chapter 1

# Introduction

## 1.1 Motivation

Federated Learning (FL) [Aba+16] has arisen as a viable alternative to conventional machine learning, enabling several clients to train a global model cooperatively without disclosing raw data. In contrast to centralized learning, which involves a server gathering and processing all data, FL preserves data on each client's device, mitigating privacy problems. Nonetheless, FL is not entirely secure, as adversaries can still deduce sensitive information from the trained model transmitted to the server. To address these privacy problems, Differential Privacy (DP) [GKN17] has been extensively included in FL to obscure individual contributions while maintaining overall model efficacy.

In FL, DP can be achieved through two main approaches: by adding noise on the client side or on the server side. In the client-side approach, each client adds noise to its local model updates before sending them to the server, ensuring that private data never leaves the device in a raw form. In the server-side approach, the server aggregates the raw updates from the clients and then applies noise to the aggregated model update, thereby masking individual contributions while updating the global model. Privacy budget ($\epsilon$) is the parameter that quantifies how much information about an individual data point may be leaked during the training process. A small $\epsilon$ means strong privacy but at the cost of lower model utility due to less accurate data. A larger $\epsilon$ means weaker privacy guarantees but a higher model utility. The amount of noise added to the model updates is based on the ($\epsilon$). A fixed privacy budget is commonly used in differential privacy (DP) to control the amount of noise added on the client side, ensuring strong privacy protection for sensitive data. However, this approach often struggles to maintain high model accuracy, especially as training progresses and the model begins to converge. This technique may result in excessive noise injection, impairing model performance. Researchers have implemented adaptive privacy budgeting algorithms to mitigate this problem. Our work is based on the *cosAFed* [Wan+24] algorithm, which employs an adaptive approach that intelligently adjusts the noise added to client updates in each round, thereby balancing privacy and accuracy. Various techniques exist for adding noise based on the required level of privacy. However, the *cosAFed* and other methods [Wei+20] [NHD20] stands out by applying noise directly to the clients' updates before transmitting them to the server.

## 1.2 Proposed Solution Approach

$SelecEval$[1] is built on the Flower framework [Beu+20] and is designed to evaluate various FL aggregation strategies using both IID and non-IID versions of the CIFAR-10 dataset [KH+09]. In an IID (independent and identically distributed) setup, each client receives a dataset that is statistically similar to the overall distribution. In contrast, in a non-IID setup, clients receive a dataset that is imbalanced, such as samples from only a subset of classes. In our case, the non-IID distribution was created using a Dirichlet quantity-based label distribution with $\alpha = 0.5$.

However, the original implementation of $SelecEval$ lacks support for DP, limiting its ability to analyze privacy-preserving FL models. This thesis aims to enhance the $SelecEval$ simulator by integrating DP algorithms and evaluating their effectiveness with existing aggregation strategies and data distributions. In particular, three DP techniques are implemented and analyzed, each offering different privacy-utility trade-offs:[2]

1. **APB-Lap**: Adaptive Privacy Budget in FL with Laplace Mechanism.

2. **APB-Gauss**: Adaptive Privacy Budget in FL with Gaussian Mechanism.

3. **APB-GAClip**: Adaptive Privacy Budget in FL with Gaussian Mechanism and Adaptive Gradient Clipping.

Each technique applies differential privacy while dynamically adapting the privacy budget. Their implementation details are as follows:

- $APB - Lap$ follows the same implementation as the $cosAFed$ [Wan+24] algorithm. It assumes that the model sensitivity is bounded by a fixed value. After updating the privacy budget, it calculates the adjustment coefficient and generates noise based on the updated privacy budget and fixed sensitivity. The Laplace mechanism [Dwo+06] is utilized to generate noise, which is then added to each client's model update before being sent to the server. By using this mechanism, we achieve ($\epsilon$)-DP guarantees, which provide stronger privacy guarantees compared to the Gaussian mechanism [DRS22] but come with several limitations.

- $APB - Gauss$ applies a similar adaptive privacy budgeting approach as $cosAfed$ but replaces the Laplace mechanism with the Gaussian mechanism. This Gaussian mechanism relies on an additional delta ($\delta$) parameter, providing ($\epsilon, \delta$)-DP guarantees. Since the Gaussian Mechanism scales noise using the $L2$ norm, it distributes noise more smoothly compared to Laplace.

- $APB - GAClip$ extends $APB - Gauss$ by incorporating adaptive gradient clipping instead of assuming a fixed sensitivity bound. This technique dynamically adjusts the clipping threshold based on the gradient distribution. By using the Gaussian mechanism, we also achieve ($\epsilon, \delta$)-DP guarantees. In this technique determining an optimal balance between noise addition, clipping sensitivity, and privacy budget is challenging.

## 1.3 Key Findings

Firstly, our results align with the $cosAFed$ algorithm [Wan+24], showing that using an adaptive $\epsilon$, as opposed to a fixed $\epsilon$, can achieve strong privacy protection without sacrificing accuracy. Secondly, another observation shows that models trained on non-IID data perform consistently better than models trained on IID data. This is because non-IID data was split using a moderate level Dirichlet distribution (with $\alpha = 0.5$), which gives each client more focused data allowing their models to specialize in fewer classes and learn the data distribution more effectively. When all client model updates are combined, this leads to a stronger global model. A third observation is that increasing the number of clients selected per training round leads to better model accuracy, which supports the findings from earlier work with *cosAFed* [Wan+24]. Fourthly, we observed that the type of above-described DP techniques has minimal influence on the evolution of the privacy budget across training rounds. This suggests that budget dynamics are more sensitive to the scoring and adjustment mechanism than the specific noise model used. Lastly, we found that the $APB - Lap$ algorithm delivered the best accuracy, followed by $APB - Gauss$ and then $APB - GAClip$. Although the Laplace mechanism is often considered less favorable than the Gaussian mechanism in theoretical literature (especially in high-dimensional settings), it delivered better performance in our setup. This result likely arises from practical factors such as the fixed sensitivity assumption and the properties of our dataset.

---
[1] https://jsincn.github.io/SelecEval/
[2] For brevity, these abbreviations will be used throughout the thesis.

# Chapter 2

# Background

## 2.1 Federated Learning

Federated Learning (FL) [Aba+16] is a decentralized machine learning technique introduced in a 2017 paper by Google. It was proposed to address growing privacy concerns and regulatory constraints surrounding centralized data collection [Aba+16]. Instead of transferring raw data to a central server, FL enables model training directly on distributed client devices, such as smartphones or edge sensors [Kai+21]. In the FL framework, local models are trained on edge devices, and only anonymized model updates are communicated to a central server. The server aggregates the model updates to update the global model.

This approach contrasts with traditional centralized machine learning, where data from multiple clients is uploaded to a single server for training. Centralized methods not only increase the risk of data breaches and misuse, but they also raise significant compliance challenges, particularly under strict privacy regulations like the General Data Protection Regulation (GDPR) in the European Union and the Health Insurance Portability and Accountability Act (HIPAA) in the United States [Rie+20].

Federated Learning addresses these issues by preserving data locality, offering stronger privacy guarantees [Bon+19], and aligning them with modern privacy-preserving computing standards. Moreover, FL supports training on highly diverse and massively distributed datasets, capturing real-world heterogeneity and potentially leading to more robust and generalizable models.

**Typical Federated Learning Workflow**

The FL process typically proceeds through the following steps:

1. **Initialization:** A central server initializes a global machine learning model with either random weights or pre-trained parameters.

2. **Client Selection:** A subset of clients is selected to participate in the current training round using Client Selection Algorithms. Such as Random, ActiveFL, and ActiveFL. The global model will be sent to selected clients.

3. **Local Training:** Each selected client trains the global model locally using private data.

4. **Model Update Upload:** After training, clients send their updated model parameters back to the central server.

5. **Server Aggregation:** The server aggregates the received updates to update the global model with different aggregation strategies, such as FedProx [Li+20], FedAvgM, and FedNova [DKM20]. This updated global model becomes the base model for the next communication round. A common method to aggregate updates is FedAvg.

6. **Iteration:** Steps 2–5 are repeated for multiple communication rounds until the global model converges.

**Privacy Challenges in Federated Learning**

Although raw data never leaves the client device in FL, the shared model updates may still leak sensitive information. Adversaries can potentially infer private attributes or even reconstruct training data using techniques like Gradient leakage, Model inversion attack, etc. For example, hackers can reverse engineer local training data from received model weights, especially when models are over-parameterized, or clients have limited data diversity. Another vulnerability lies in how updates are sent over networks. If these exchanges aren't properly secured, the communication of unprotected updates across networks opens up possibilities for man-in-the-middle attacks. Thus, while FL reduces direct data sharing, it still requires robust privacy-preserving mechanisms to ensure end-to-end confidentiality in sensitive applications such as healthcare, finance, and personalized services.

## 2.2 Differential Privacy

To address the privacy problem in FL, Differential Privacy (DP) [DR+14] is one of the techniques to protect sensitive data. DP is a mathematical framework that quantifies the privacy guarantees of a computation. It ensures that the inclusion or exclusion of a single data point in the training set has a minimal impact on the output of the algorithm.

**Formal Definition of DP**

A randomized mechanism $M$ is said to be $(\epsilon, \delta)$-differentially private if, for all datasets $D_1$ and $D_2$ differing by one record, and for all outputs $O$:

$$\Pr[M(D_1) \in O] \leq e^\epsilon \cdot \Pr[M(D_2) \in O] + \delta \tag{2.1}$$

- $\epsilon$ (epsilon) – the privacy budget, which controls the privacy-utility tradeoff.

- $\delta$ (delta) – the probability that the privacy guarantee may fail.

- $M$ (a randomized mechanism) - like a federated learning process.

It says that the probability that a randomized algorithm $M$ outputs results in some set $O$ when run on dataset $D_1$ is at most $e^\epsilon$ times the probability that it outputs results in $O$ when run on a neighboring dataset $D_2$, plus a small value $\delta$. Where $D_1$ and $D_2$ are called neighboring datasets, meaning they differ by only one individual's data.

**Informal Explanation of DP**

Suppose we have a trained machine learning model *M* that predicts whether a person has cancer using health data. This model is trained on a dataset *D*, which includes many users' medical records. Let's say that two versions of this dataset:

- $D_1$ A dataset that includes user Rakesh's data.

- $D_2$ The same dataset, but without Rakesh's data

These two datasets $D_1$ and $D_2$ are neighboring datasets because they differ by only one person's data.

We use the model *M* trained on $D_1$, and it says, "There is an 80% chance that Rakesh has cancer". Now we remove Rakesh's data from the dataset and train a new model on $D_2$, and it says, "There is a 55% chance that Rakesh has cancer." Even though we changed the model with only one data point, we get 25% different. It means that the model output depends greatly on Rakesh's data, which is a privacy risk.

A DP algorithm makes sure that such differences are very small. If the model satisfies $(\epsilon, \delta)$-DP, then even if someone knows the output, they cannot confidently tell whether Rakesh's data was used or not. In practice, it might mean both outputs (with and without Rakesh) would be around 77% and 75%, so the attacker cannot really tell whether Rakesh's data was included.

## 2.3 Applying Differential Privacy in Federated Learning

In FL, to make the training process differentially private, we must ensure that the server cannot infer sensitive information from the updates it receives from clients. To achieve that, noise can be added in two ways:

- **Client-side noise addition**: Each client adds noise to its model updates before sending them to the server.

- **Server-side noise addition**: The server adds noise to the aggregated updates before updating the global model.

In our work, we adopt the **client-side noise addition** approach, as it is also employed by our base paper [Wan+24]. To ensure differential privacy during the Federated Learning process, we utilize the following two mechanisms to inject noise on client-side.

**1. Laplace Mechanism**

The Laplace mechanism [Dwo+06] is used when satisfying **pure differential privacy** (i.e., $\delta = 0$). Given a function $f : \mathcal{D} \to \mathbb{R}^k$, the Laplace mechanism adds noise drawn from a Laplace distribution:

$$M(D) = f(D) + \text{Lap}\left(\frac{\Delta f}{\epsilon}\right) \qquad (2.2)$$

where *M* is the mechanism that compute the function *f* and adds Laplace noise to its output *O*, and $\Delta f$ is the **sensitivity** of *f*, defined as:

$$\Delta f_1 = \max_{D_1, D_2} \|f(D_1) - f(D_2)\|_1 \qquad (2.3)$$

## 2. Gaussian Mechanism

The Gaussian mechanism [DRS22] is used when we allow a small failure probability ($\delta > 0$), making it suitable for ($\epsilon, \delta$)-DP. It is less private, as we use the failure probability $\delta$. Given a function $f : \mathcal{D} \to \mathbb{R}^k$, the Gaussian mechanism adds noise from a Gaussian distribution:

$$M(D) = f(D) + \mathcal{N}\left(0, \sigma^2 I\right) \tag{2.4}$$

where $\sigma$ (standard deviation of the noise) is chosen based on the $\ell_2$ **sensitivity**:

$$\Delta_2 f = \max_{D_1, D_2} \|f(D_1) - f(D_2)\|_2 \tag{2.5}$$

To ensure ($\epsilon, \delta$)-DP, the noise scale $\sigma$ must satisfy:

$$\sigma \geq \frac{\Delta_2 f \cdot \sqrt{2\ln(1.25/\delta)}}{\epsilon} \tag{2.6}$$

These mechanisms ensure that no single user or client has a noticeable effect on the trained model, thus preserving user privacy during federated learning. In our work, $APB-Lap$ uses the Laplace Mechanism to add the noise on the client side before sending updates to the server. Also, $APB-Gauss$ and $APB-GAClip$ use the Gaussian mechanism to add the noise on the client side before sending its updates to the server.

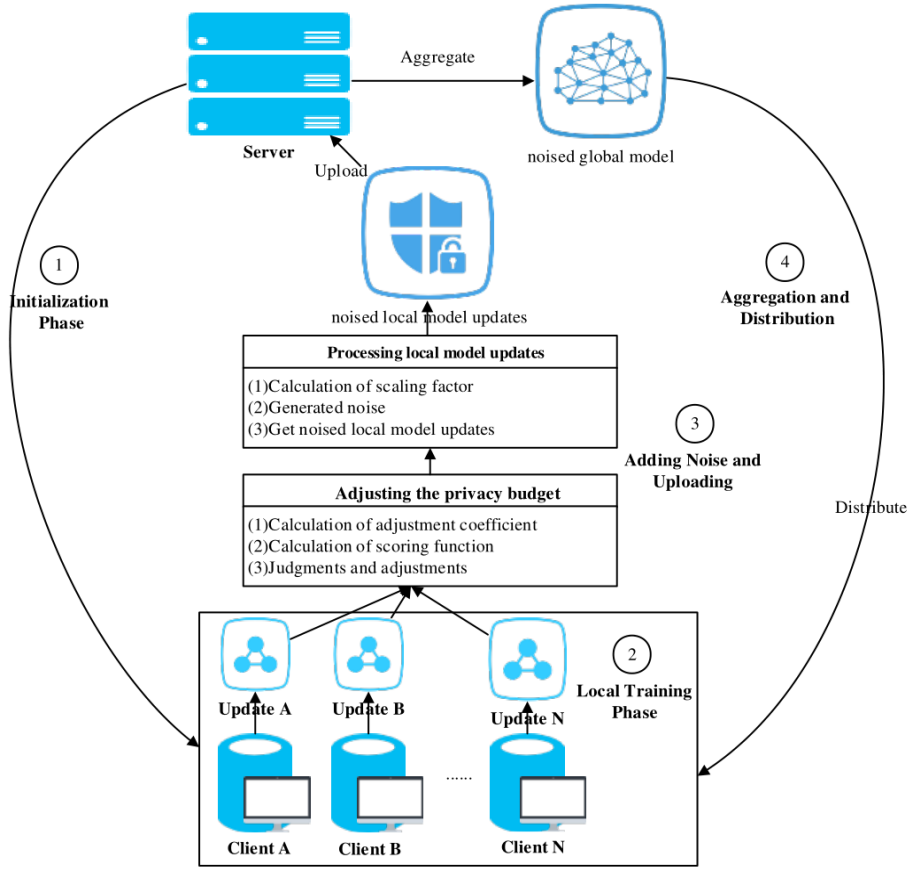

**Figure 2.1:** Process of applying noise to the model update based on cosAFed [Wan+24]

## 2.4 Gradient Clipping and Sensitivity Control

As you can see from the Formals 2.4 and Formal 2.2, adding noise is based on the Sensitivity $\Delta f$. To apply DP effectively, it is necessary to bound the sensitivity of the model updates, which refers to the maximum change that one data point can cause in the output [Liu24]. This is usually done through gradient clipping.

- **Fixed Clipping**: A constant threshold (e.g., 1.0) is applied to clip gradients before adding noise.

- **Adaptive Clipping**: The clipping threshold is dynamically updated in each round, typically using the 90th percentile or exponential moving average of the observed gradient norms.

In $APB-Lap$ and $APB-Gauss$, we assume that $\Delta f$ is already bounded to the fixed value. So we do not need to clip the gradients. Whereas in $APB-GAClip$, we do not assume the sensitivity, so we clip the updates gradients adaptively. The threshold is initially defined before the training starts. In our work, we take this value as $\Delta f = 5$. During the round, it is dynamically updated. This threshold will clip the norm of gradient updates to the initial decided clipping value. Hence, we get the sensitivity $\Delta f$ as the initially defined clipping value during the noise addition. This makes the FL process differentially private. The process of adding the noise illustrated in Figure 2.1, and this technique is used in all three DP algorithms.

## 2.5 *SelecEval* Framework

*SelecEval*[1] is a simulator built on top of the Flower framework that allows researchers to compare different FL aggregation strategies (e.g., FedAvg, FedProx, FedNova, FedDisco) and client selection algorithms. Originally, *SelecEval* did not support differential privacy, limiting its applicability for privacy-focused experiments.

In this thesis, we extend *SelecEval* by integrating support for differential privacy mechanisms and privacy budget adaptation. Specifically, we apply DP to FedAvg and FedProx strategies by modifying the client and server logic. Other aggregation strategies such as FedAvgM and FedNova were not adapted due to their use of parameter buffers, which are incompatible with update-level noise injection.

## 2.6 Related Work

DP has become a core component in FL to mitigate the risk of information leakage through shared model updates. Two primary approaches exist: client-side (local) DP and server-side (central) DP.

In client-side DP, noise is added directly to each client's model update before transmission. This offers strong privacy even against untrusted servers. Local DP in FL was introduced in earlier works [GKN17], and the "Noising before Aggregation" (NbAFL) strategy was later proposed to ensure user-level privacy by injecting noise at the client level [Wei+20]. However, local DP often requires large noise, which may degrade accuracy especially in high-dimensional models. To address this, various enhancements such as adaptive clipping [And+21] and selective update sharing [Tru+20] have been proposed to reduce this effect.

---
[1] https://jsincn.github.io/SelecEval/

In server-side DP, noise is added after aggregation. Methods like DP-FedAvg [Aba+16] [McM+17] assume a semi-trusted server and apply DP-SGD-style clipping and noise injection to the aggregated updates. This reduces the required noise per client and preserves higher model accuracy, making it suitable for large-scale deployments such as Google's mobile keyboard models [McM+22].

Regarding DP mechanisms, the Laplace mechanism provides $(\epsilon, 0)$-DP using $L_1$ sensitivity, while the Gaussian mechanism supports $(\epsilon, \delta)$-DP using $L_2$ sensitivity [DR+14]. Most FL studies prefer Gaussian noise due to better performance in iterative and high-dimensional settings [Tru+20], although Laplace noise may be used when stronger privacy guarantees are needed. A comparative study between the two mechanisms shows that Gaussian noise generally results in faster convergence and higher accuracy under the same privacy budget $\epsilon$ [Jin+23].

In summary, both client-side and server-side DP approaches are viable. Client-side DP offers stronger trust guarantees but may harm utility. Server-side DP assumes some trust in the aggregator but typically results in better performance. The choice of mechanism, noise calibration, and clipping strategies significantly impact the privacy-utility trade-off in FL.

# Chapter 3

# Implementation

## 3.1 Overview of Implementation

We have extended the *SelecEval* simulator by incorporating DP algorithms. In our implementation, we integrate three types of DP algorithms. The existing *SelecEval* framework already supports five different aggregation strategies: FedNova, FedAvg, FedAvgM, FedProx, and FedDisco. These strategies, along with model compression techniques, were introduced in previous research. Their primary role is to aggregate model parameters that clients send to the server in each round.

Typically, in DP-based FL, noise is added to model updates rather than directly to model parameters. However, the existing implementation aggregates model parameters (Weights of the neural network after model is trained) instead of model updates (different between global model parameters and trained model parameters). This is not a common practice in FL with DP since it can significantly degrade model performance.

Initially, we followed the approach of adding noise to model parameters and then applying existing aggregation strategies to model parameters. However, as our results were unsatisfactory, we modified the implementation to aggregate model updates instead of model parameters. Specifically, we adapted FedProx and FedAvg, modifying them to work with model updates rather than model parameters.

We did not modify FedAvgM, FedDisco, or the original version of FedNova, as these strategies rely on momentum buffers for model parameters on the server side. This restriction prevents us from sending only model updates. Therefore, we focused on integrating DP with FedProx and FedAvg.

### 3.1.1 APB-Lap

Our extension of *SelecEval* is based on the *cosAFed* [Wan+24] differential privacy (DP) algorithm. A key feature of *cosAFed* is its adaptive privacy budget mechanism, which dynamically adjusts privacy parameters in each round. The algorithm consists of three main steps:

1. **Computation of Adjustment Coefficient and Scoring Function** After receiving the global model, clients train their local models using their respective datasets. The *cosAFed* algorithm then computes:

    - The **adjustment coefficient**, which is determined based on the cosine similarity between the local and global models from the previous round, the number of datasets, the number of client datasets, the total number of clients, and the number of selected clients.

- The **scoring function**, which is calculated using the accuracy and loss of the global model from previous rounds, as well as the total number of training rounds.

Based on these values, the privacy budget is adjusted dynamically.

2. **Computation of Scaling Factor and Noise Generation** The **scaling factor** is computed using the gradient norms of the global model from the previous round and the local model. Noise is then generated based on the updated privacy budget and the local model update.

3. **Noise Addition and Model Aggregation** The computed **scaling factor** and **generated noise** are applied to the local model before sending it to the server. The **noised model updates** are then aggregated at the server to form the new global model.

*cosAFed* maintains flexibility in privacy settings during the initial stages of training, allowing the model to learn effectively and improve accuracy. If a client's updates are similar to previous ones, additional noise is introduced to enhance privacy as the privacy budget decreases. Conversely, if the client's updates show significant differences, the privacy budget is reset to its initial value, resulting in less noise being added. The grading system determines whether noise should be added. The pseudo-code detailing this process is presented in Algorithm 1 & 2.

---

**Algorithm 1** *cosAFed* [Wan+24]

**Input:** epochs $T$, initial global model $A^0$, clients $C = \{C_1, C_2, \ldots, C_M\}$, number of clients per round $N$, datasets $D = \{D_1, D_2, \ldots, D_M\}$, learning rate $\eta$, initial privacy budget $\varepsilon$, assumed sensitivity of the model $\Delta f$
**Output** the noised global model $A^T$
$B_1^0, B_2^0, \ldots, B_N^0 \to A^0$
**Repeat**
    **for** ($t \to 1; t \leq T; t \to t+1$)
        **Select** $N$ clients out of $M$ and **send** $\{A^{t-1}\}$
        **for** ($i \to 1; i \leq N; i \to i+1$)
            $B_i^t \to \mathbf{SGD}(A_i^{t-1}, \eta, D_i)$
            $\Delta B_i^t \to A_i^{t-1} - B_i^t$
            $\varepsilon_i^t \to \mathbf{ADJUST}(M, N, D_i, D, B_i^t, A_i^{t-1}, t, T, \text{acc}, \text{loss}, \varepsilon)$
            $\phi \to \dfrac{A_i^{t-1} \cdot B_i^t}{\|A_i^{t-1}\| \|B_i^t\|}$
            noisy $\to \mathbf{L}(0, \Delta f / \varepsilon_i^t)$
            $Q_i^t \to \phi \cdot \Delta B_i^t + \text{noisy}$
            **Upload** $Q_i^t$
        **End for**
        $A^t \to \mathbf{FEDPROX}(Q^t, A^{t-1})$
        $\mathbf{SAVE}(\text{acc}, \text{loss})$
    **End for**
**Until** $t = T$
**return** $A^T$

---

**Algorithm 2** $ADJUST(arg1, arg2, ...)$ [Wan+24]

**Input:** epochs $T$, current round $t$, global model $A_i^{t-1}$, local model of client $i$ in round $t$ $B_i^t$, local model update $\Delta B_i^t$, clients $C = \{C_1, C_2, ..., C_M\}$, number of participating clients $N$, privacy budget $\varepsilon$, datasets $D = \{D_1, D_2, ..., D_M\}$, accuracy history $\text{acc} = \{\text{acc}_1, \text{acc}_2, ..., \text{acc}_{t-1}\}$, loss history $\text{loss} = \{\text{loss}_1, \text{loss}_2, ..., \text{loss}_{t-1}\}$
**Output:** adjusted privacy budget $\varepsilon_i'$
**Compute scaling factor**
$$S_c \to \frac{A_i^{t-1} \cdot B_i^t}{\|A_i^{t-1}\|\|B_i^t\|}$$
**Compute adjustment coefficient** $p$
    **if** $S_c \geq 0$
        $p \to |1 - (S_c \times M \times |D_i|)/(N \times |D|)|$
    **else**
        $p \to 1$
**Compute scoring function**
    **if** $\text{loss}_{t-1} \geq \text{loss}_{t-2}$
        $\text{score}_{\text{loss}} \to 1$
    **else**
        $\text{score}_{\text{loss}} \to 0$
    $\text{tempAcc} \to 0$
    **for** $(o \to 0; o < N; o \to o+1)$
        **if** $t - 1 - N + o < 0$
            $\text{tempAcc} \to \text{tempAcc}$
        **else**
            $\text{tempAcc} \to \text{tempAcc} + \text{acc}_{t-1-N+o}$
    **end for**
    **if** $\text{tempAcc} \geq \text{acc}_{t-1}$
        $\text{score}_{\text{acc}} \to 1$
    **else**
        $\text{score}_{\text{acc}} \to 0$
    **if** $t/T \geq 1/2$
        $\text{score}_t \to 1$
    **else**
        $\text{score}_t \to 2 \times t/T$
    $\text{score} \to 30 \times \text{score}_{\text{loss}} + 40 \times \text{score}_{\text{acc}} + 30 \times \text{score}_t$
**Adjust privacy budget**
    **if** $(\text{score} > 50)$ **and** $(p \leq 1)$
        $\varepsilon' \to p \times \varepsilon$
    **else**
        $\varepsilon' \to \varepsilon$
**return** $\varepsilon'$

---

### 3.1.2 APB-Gauss

In this modification, we adopt a similar approach to *APB-Lap*, but instead of using the Laplace Noise Mechanism, we apply the Gaussian Mechanism. The gradient norms are not clipped; instead, we keep them fixed, following the same principle outlined in [Wan+24]. Consequently, the sensitivity value remains the same as in the Laplace mechanism and is fixed.

    It's important to note that the Gaussian Mechanism introduces noise based on a parameter

$\delta$, which is chosen according to the total number of clients, as described in [Pon+23]. In our experiments, we set $\delta = 0.01$ for a total of 100 clients. The pseudo-code for this algorithm is obtained by replacing the *Laplace noise* in Algorithm 1 with Gaussian noise, as described in the format below.

$$\text{noisy} \to \mathcal{N}\left(0, \frac{\Delta f^2 \cdot \log(1.25/\delta)}{\varepsilon_i t^2}\right) \tag{3.1}$$

### 3.1.3 APB-GAClip

In our implementation, we incorporate the concept of adaptive clipping, which plays a crucial role in FL with DP. Adaptive clipping dynamically adjusts the clipping threshold of model gradients in each round based on the values observed from the selected clients in the previous round. Specifically, we use a quintile-based approach [And+21] to determine the new clipping value: the server collects all noisy model updates from the selected clients, computes the L2 norm of each update, and then sets the clipping threshold for the next round based on the 90th percentile of these norms.

For the initial round, since no prior information is available, we set the clipping value to 5. This is a good starting point, allowing the clipping threshold to increase during early training and decrease later as the model converges, helping improve final accuracy. To update the clipping threshold smoothly over time, we apply the Exponential Moving Average (EMA) method, as described in [Can+21]. The updated threshold is then shared with the clients participating in the next round. Each client clips their model update using this new threshold.

When adding Gaussian noise to the clipped updates, we again apply EMA to the sensitivity value. This helps stabilize the training process, as experiments show that large fluctuations in sensitivity can lead to unstable model behavior. The pseudo-code for this process is provided below, incorporating two learning rates: one for updating the clipping threshold and one for updating the sensitivity. This approach also builds on Algorithm 2, which manages the privacy budget throughout the training.

## 3.2 Modifications to *SelecEval* Simulator

### 3.2.1 Client-Side DP Integration

To integrate the Differential Privacy (DP) mechanism into *SelecEval*, we modified both the client and server files. On the client side, after training the model, the privacy budget is updated based on the previous value. Then, noise is added to each client's model update before being sent to the server for aggregation.

For the Gaussian mechanism, in each round, clients receive an updated clipping value from the server. This value is used to clip the model update gradients before adding noise.

In *SelecEval*, clients are generated automatically, with each one having its own separate object. Once a client sends its model update to the server, its object is destroyed. This allows the system to work in parallel, but before aggregation, the server ensures that it has received updates from all clients within a specified time frame.

---

**Algorithm 3** cosAFed with Gaussian Noise and Threshold

**Input:** epochs $T$, initial global model $A^0$, clients $C = \{C_1, C_2, \ldots, C_M\}$, number of clients per round $N$, datasets $D = \{D_1, D_2, \ldots, D_M\}$, learning rate $\eta$, initial privacy budget $\varepsilon$, initial threshold $\theta^0$, lr_sensitivity $\zeta$, lr_threshold $\tau$
**Output** the noised global model $A^T$
$B_1^0, B_2^0, \ldots, B_N^0 \rightarrow A^0$
**Repeat**
    **for** ($t \rightarrow 1; t \leq T; t \rightarrow t+1$)
        **Select** $N$ clients out of $M$ and **send** $\{A^{t-1}, \theta^t\}$
        **for** ($i \rightarrow 1; i \leq N; i \rightarrow i+1$)
            $B_i^t \rightarrow \text{SGD}(A_i^{t-1}, \eta, D_i)$
            $\Delta B_i^t \rightarrow A_i^{t-1} - B_i^t$
            $\varepsilon_i^t \rightarrow \text{ADJUST}(M, N, D_i, D, B_i^t, A_i^{t-1}, t, T, \text{acc}, \text{loss}, \varepsilon)$
            $\Delta B_i'^t \rightarrow \text{CLIP}(\Delta B_i^t, \theta^t)$               ▷ use L2 Norm
            $\phi \rightarrow \dfrac{A_i^{t-1} \cdot B_i^t}{\|A_i^{t-1}\| \|B_i^t\|}$
            $\Delta f' \rightarrow \text{updateDelta}(\Delta f, \zeta, \theta^t)$         ▷ use EMA method
            noisy $\rightarrow \mathcal{N}(0, \dfrac{\Delta f'^2 \cdot \log(1.25/\delta)}{\varepsilon_i t^2})$
            $Q_i^t \rightarrow \phi \cdot \Delta B_i'^t + \text{noisy}$
            **Upload** $Q_i^t$
        **End for**
        $A^t \rightarrow \text{FEDPROX}(Q^t, A^{t-1})$
        $\theta^t \rightarrow \text{UPDATETHRESHOLD}(\theta^{t-1}, \tau, t, \ldots)$     ▷ EMA on the serverside
        SAVE(acc, loss)
    **End for**
**Until** $t = T$
**return** $A^T$

---

### 3.2.2 Server-Side Aggregation

We modified FedProx and FedAvg in the existing *SelecEval* simulator. As mentioned earlier, the original aggregation strategies in *SelecEval* aggregate model parameters rather than model updates.

On the server side, before performing aggregation, we calculate a new clipping value for the next round. This value is determined based on the L2 norm of the model updates received from the selected clients. Specifically, we use the 90th percentile of these norms to set the clipping value, which is then applied in the second round.

Other aggregation strategies—FedNova, FedDisco, and FedAvgM—cannot be modified for differential privacy. These strategies rely on buffer-based aggregation, where the server maintains a buffer of model parameters. Since DP requires modifying model updates rather than model parameters, we kept these three strategies unchanged.

## 3.3 Experimental Setup

We conducted a series of experiments using the CIFAR-10 dataset to evaluate the performance of different DP algorithms under varying conditions. The total number of clients was set to 100, and each experiment was run for 200 communication rounds. In every round, a subset of clients was randomly selected to participate. In all experiments, either the FedProx [Li+20] or FedAvg [McM+17] aggregation strategy is selected. The common configuration parameters for these experiments are summarized in Table 3.1, while algorithm-specific parameters are detailed in Table 3.2.

| Parameter | Value |
|---|---|
| no_rounds | {200, 300} |
| algorithm (CS) | random |
| dataset | cifar10 |
| no_epochs | 2 |
| var_epochs | false |
| total_clients | 100 |
| selected_clients | {10, 20} |
| batch_size | 32 |
| timeout | 200 |
| **data_config** | |
| data_quantity_skew | Dirichlet, Uniform |
| data_label_skew | Dirichlet, Uniform |
| data_label_distribution_parameter | 0.5 |
| data_quantity_distribution_parameter | 0.5 |
| data_quantity_min_parameter | 32 |
| data_quantity_max_parameter | 2000 |
| data_feature_skew | None |
| **simulation_config** | |
| network_bandwidth_mean | 20.0 |
| network_bandwidth_std | 10.0 |
| network_bandwidth_min | 0.0 |
| performance_factor_mean | 1.0 |
| performance_factor_std | 0.2 |
| reliability_parameter | 1.0 |
| create_synthetic_client_failures | true |

**Table 3.1:** Common Configuration Parameters

### 3.3.1 Baseline Model

To compare the impact of differential privacy mechanisms, we also trained a baseline model without adding any noise, i.e., without applying DP. This setup used the same number of clients, rounds, and random selection strategy as the DP experiments to ensure a fair comparison.

---

[1]Relevant for APB_Lap and ABP_Gauss: A configuration with Fixed Sensitivity = 0.5 was tested using $\epsilon = 150$ and IID data with FedProx Only.

| Parameter | Value |
|---|---|
| **APB_Lap_config** | |
| epsilon($\epsilon$) | {100, 120, 150, 170, 200} |
| fixed_sensitivity($\Delta f$) | {0.5, 1.0} |
| aggregation_strategy | {FedProx, FedAvg} |
| **APB_Gauss_config** | |
| epsilon($\epsilon$) | {100, 120, 150, 170, 200} |
| delta($\delta$) | 0.01 |
| fixed_sensitivity($\Delta f$) | {0.5, 1.0} |
| aggregation_strategy | FedProx |
| **APB_GAClip_config** | |
| epsilon($\epsilon$) | {100, 120, 150, 170, 200} |
| delta($\delta$) | 0.01 |
| lr_sensitivity | 0.0001 |
| lr_threshold | 0.5 |
| initial_threshold | 5 |
| aggregation_strategy | FedProx |

**Table 3.2:** Algorithm Configuration Parameters[1]

**Dataset**

We use the CIFAR-10 [KH+09] dataset, a standard benchmark in image classification and federated learning research. It consists of 60000 color images across 10 classes, with 50000 training samples and 10000 test samples. Both IID and non-IID versions of the dataset are used in the experiments. The non-IID partitioning uses a Dirichlet distribution with $\alpha = 0.5$ for both label and quantity distributions, allowing clients to specialize in fewer classes.

### 3.3.2 FL Aggregation Strategies

Two existing aggregation strategies are used and compared:

- **FedAvg [McM+17]:** A widely-used baseline in FL, where the global model is obtained by averaging the updates from all participating clients in each round. While simple and effective under IID data, FedAvg can perform poorly with non-IID client distributions.

- **FedProx: [Li+20]** An extension of FedAvg that introduces a proximal term to penalize local updates that deviate too far from the global model. This strategy is particularly useful in heterogeneous environments with unbalanced data distributions.

### 3.3.3 Privacy Budgets and DP Mechanisms

To evaluate the trade-off between privacy and utility, we experiment with a range of privacy budgets $\epsilon \in \{100, 120, 150, 170, 200\}$ while keeping the delta value fixed at $\delta = 0.01$. Privacy Budgets values were chosen based on previous work, such as [Wan+24], which showed meaningful variation in model performance across this range. Delta value is chosen based on the number of clients in the training according to the paper [Pon+23]. We also include a baseline experiment with no differential privacy applied (i.e., without noise).

Three differentially private mechanisms are used in the experiments:

- **Gaussian Mechanism:** Noise is added assuming a fixed clipping threshold, with values of 1.0 or 0.5.

- **Laplace Mechanism:** Similar to the Gaussian mechanism but using Laplace noise. Experiments are conducted with fixed clipping thresholds of 1.0 and 0.5.

- **Adaptive Clipping with Gaussian Mechanism:** Here, the clipping threshold is dynamically updated using a learning rate of 0.5, with sensitivity updated at a learning rate of 0.0001. The initial clipping threshold is set to 5.0.

### 3.3.4 Client Selection Strategy

Each round involves a random [McM+17] selection of clients from a pool of 100. Depending on the configuration, either 10 or 20 clients are selected per round. This randomized approach allows us to observe how increasing the number of selected clients affects model accuracy and convergence, as well as the behavior of the privacy budget.

### 3.3.5 Training Rounds and Convergence

Although initial experiments followed the base paper's setup of 300 rounds, we found that the model achieved approximately 80% accuracy within the first 100 rounds. To optimize runtime and better observe convergence trends, we fixed the number of communication rounds at 200 for all experiments. Each round includes two local training epochs.

### 3.3.6 Clipping Threshold

We evaluate the global model using both Fixed Clipping and Adaptive Clipping thresholds. Clipping is an essential technique in differential privacy, ensuring that individual client updates remain within a bounded range before noise is applied. This prevents any single client from having an outsized influence on the global model while preserving privacy guarantees.

- Fixed Clipping Threshold: The fixed threshold value is taken from the base paper, which uses 1.0. However, in our experiments, we do not explicitly clip the gradient updates using this fixed threshold. Instead, we assume that our local model updates are naturally bounded by this fixed value. This approach aligns with the standard practice in differential privacy for FL settings in Paper [Wan+24].

- Adaptive Clipping Threshold: This method is applied when implementing the Gaussian mechanism. Unlike the fixed threshold, adaptive clipping is more complex, as small changes in the threshold can significantly impact model accuracy. In our experiments, we initialize the clipping value at 1.0 and 5.0 based on observations from prior experimental results. Choosing an excessively high value can cause $NaN$ (Not a Number) errors in model accuracy, as adaptive clipping directly influences the amount of noise added to the updates. If too much noise is introduced in each round, it can result in an unstable and unbalanced global model, negatively affecting the next training rounds.

# Chapter 4

# Experimental Results

All experiments were performed on a workstation running **Ubuntu 20.04**, equipped with an **AMD Ryzen 9 5950X CPU** (32 cores), **128 GB RAM**, and an **NVIDIA RTX 3060 GPU** with 12 GB of VRAM.

## 4.1 Evaluation of the APB-Lap Algorithm

### 4.1.1 Effect of FedProx and FedAvg on Model Accuracy

To evaluate the impact of the APB-Lap, we compared the accuracy of FedAvg and FedProx aggregation strategies over 300 training rounds. A range of privacy budget values was used, assuming the gradient norms of updates are bounded to 1.0 when applying Laplace noise. This assumption holds across all experiments in the following subsections. The experiments were conducted on both IID and Non-IID versions of the validation dataset across clients, and 10 clients were selected in each round.

As shown in Figure 4.1 and Figure 4.2, FedAvg and FedProx produced nearly identical results across all privacy budgets in both IID and Non-IID settings. This is consistent with previous findings [Wür24], which also reported similar performance between the two strategies under comparable conditions.

Furthermore, we observe that as the privacy budget increases toward infinity, the results gradually approximate the accuracy of the non-private model (NoDP). This aligns with the findings from paper [Aba+16] on DP in FL, confirming that our experimental results are consistent with established literature.

### 4.1.2 Effect of IID and Non-IID Datasets on Model Accuracy

Initially, experiments were conducted for 300 training rounds using the IID dataset. However, as shown in Figure 4.1, validation accuracy plateaued between rounds 200 and 300, showing no significant improvement across clients. For efficiency, we reduced the number of training rounds to 200 in subsequent experiments. As the FedProx and FedAvg achieve similar results, also found in paper [Wür24], we conducted further experiments using FedProx, as it includes a proximal term that helps stabilize updates for Non-IID datasets.

Figure 4.3, shows the mean validation accuracy with the FedProx aggregation strategy applied to different data distribution. We use privacy budget values as {100, 200}, and 10 clients are selected in each round.

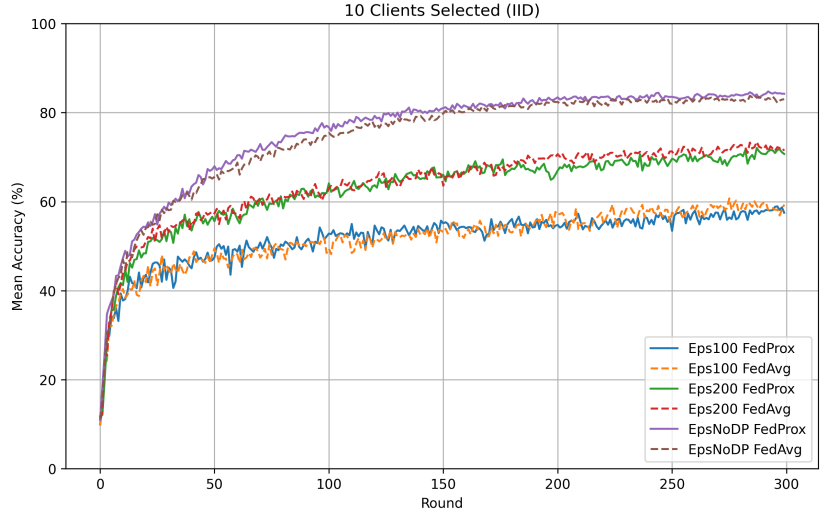

**Figure 4.1:** Effect of FedProx and FedAvg on model accuracy with IID dataset

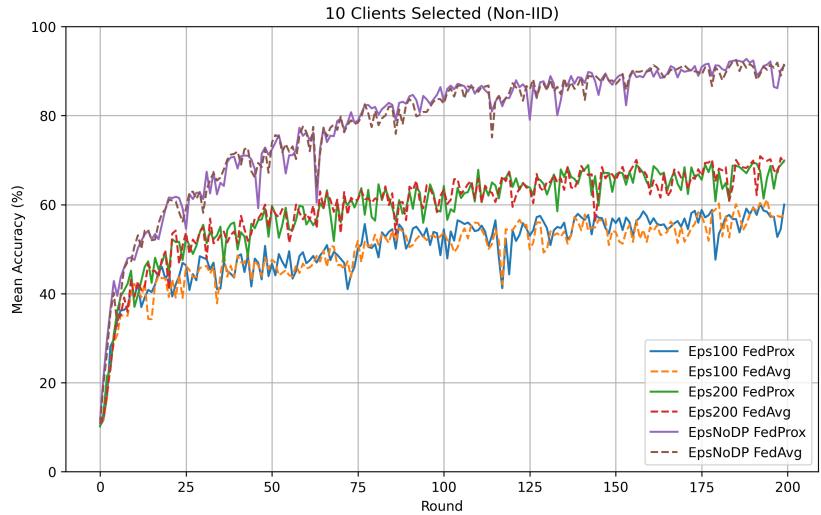

**Figure 4.2:** Effect of FedProx and FedAvg on model accuracy with Non-IID dataset

We can observe that with a privacy budget, 100 shows lower accuracy and more fluctuation, indicating the stronger impact of noise. While increasing the privacy budget to 200 improved both accuracy and stability. These results align with the paper [Wan+24].

When we compare the Baseline model (Without adding noise), the Non-IID dataset achieves higher accuracy, peaking close to 85%, followed by the IID dataset, which stabilizes around 78%. This is because the baseline model shows the upper performance bound in the absence of noise, confirming that DP introduces a trade-off between accuracy and privacy.

However, when we add the noise with privacy budgets 100 and 200, it shows that the IID dataset achieves slightly better accuracy compared to the Non-IID dataset. This is because the Non-IID dataset is more sensitive to the noise, resulting in unstable model updates and higher variability.

When we compare the DP model with the baseline model, the Non-IID dataset achieves higher accuracy than the IID dataset. It is unlikely to get this, but we use moderate level Dirichlet quantity and label distribution with $\alpha$ value 0.5. Each client receives the particulate label class with a high amount of data, which leads to better local model accuracy. Comparing IID vs. Non-IID behavior, we observe that IID models (with and without DP) show smoother accuracy curves, while Non-IID models exhibit greater fluctuation and struggle to generalize well. This is due to inconsistent local updates and a lack of class diversity in non-IID datasets.

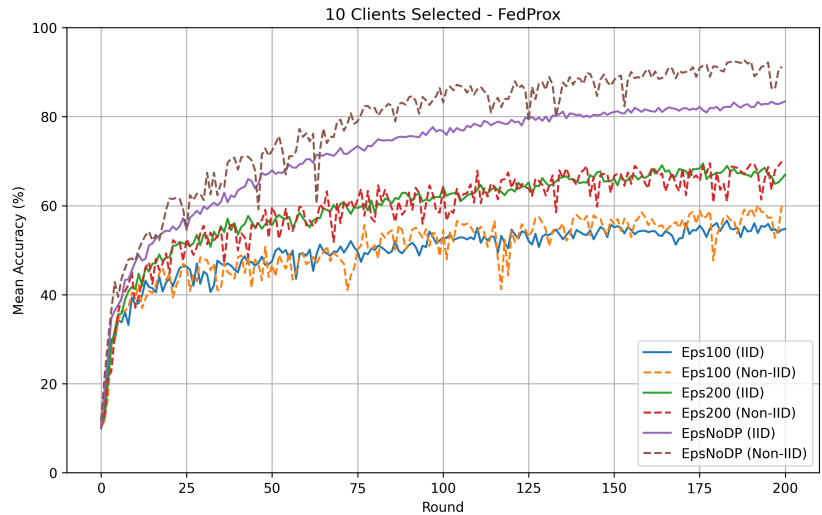

**Figure 4.3:** Effect of IID and Non-IID datasets on model accuracy

### 4.1.3 Effect of Privacy Budget on Final Round Accuracy

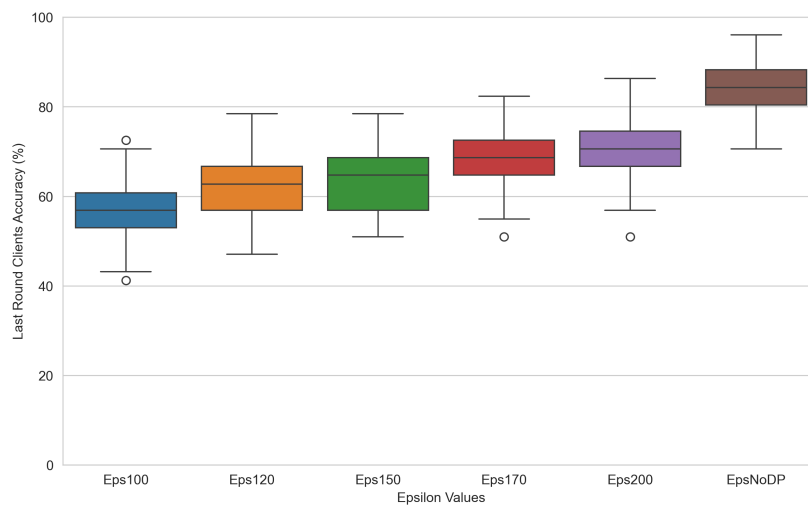

**Figure 4.4:** Effect of privacy budget on final round accuracy

As depicted in Figure 4.4, the model's accuracy is directly influenced by the chosen privacy budget. A higher privacy budget allows the model to learn more quickly but offers weaker privacy protection. Consequently, as the budget increases incrementally, the final accuracy likewise improves. Notably, when no noise is applied, the model reaches its highest accuracy at 90%.

### 4.1.4 Effect of Numbers of Selected Clients on Model Accuracy

We conducted an experiment using varying numbers of selected clients per round, with a fixed privacy budget of 200, on an IID dataset. The training process was divided into five equal segments of 40 rounds each. For every segment, we recorded the highest validation accuracy achieved. On the X-axis, **G1_1st** corresponds to rounds 0–40, where the best accuracy was 57.08% with 10 selected clients per round. **G2_1st** represents the next 40 rounds (41–80), and this pattern continues up to 200 rounds. Similarly, **G1_2nd** refers to the first 40 rounds (0–40) when 20 clients were selected in each round.

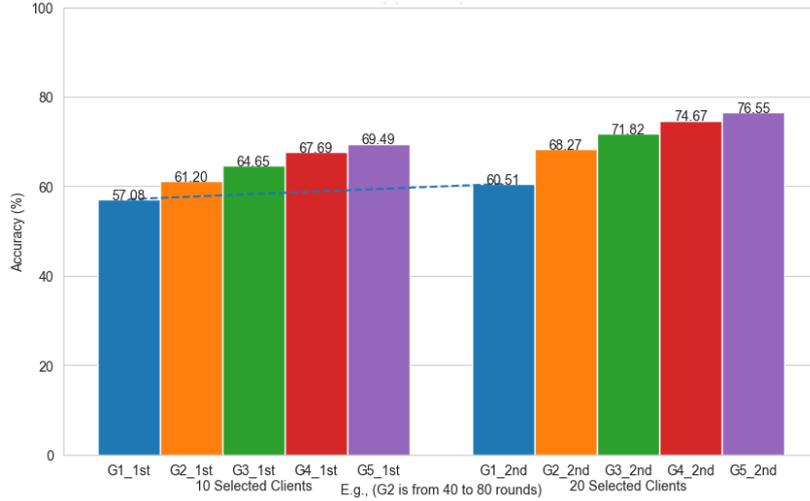

**Figure 4.5:** Max validation accuracy per group with 10 or 20 selected clients

As shown in Figure 4.5, accuracy improves steadily from G1 to G5 in both setups, indicating effective convergence. Across all groups, models with 20 selected clients consistently outperform those with 10 clients, achieving higher accuracy. For example, accuracy increases from 57.08% in **G1_1st** to 60.51% in **G1_2nd**, as highlighted by the dotted blue line. A similar trend is observed in other groups—for instance, from **G3_1st** to **G3_2nd**, where accuracy also improves. We can see this trend in all the groups of rounds.

This performance gain can be attributed to two main factors. First, selecting more clients per round increases the total data volume, enabling the model to better capture the global data distribution. Second, a higher number of selected clients improves the adjustment coefficient, which effectively increases the usable privacy budget. This results in less noise being added during training, thereby enhancing the overall accuracy.

### 4.1.5 Effect of Number of Selected Clients on Updated Privacy Budget

Figure 4.6 shows the effect of a number of selected clients on the updated privacy budget under the APB-Lap mechanism, using a fixed target budget (100, 120, 150, 170, and 200) for each configuration. In each round, 10 or 20 clients are selected, and IID datasets are used.

Two key observations can be made. First, the range of changes in the privacy budget increases as the initial value of the privacy budget increases. For example, with an initial privacy budget of 200, the observed change is larger ($200 - 180 = 20$) compared to an initial privacy budget of 120 ($120 - 110 = 10$). This trend is evident in both client selection setups. The reason lies in the update mechanism of the privacy budget: when the adjustment coefficient $p < 1$ (as described in Algorithm 2), the updated privacy budget $\epsilon'$ is computed by multiplying the previous value with $p$. Consequently, higher initial privacy budgets lead to greater absolute changes compared to lower initial budgets.

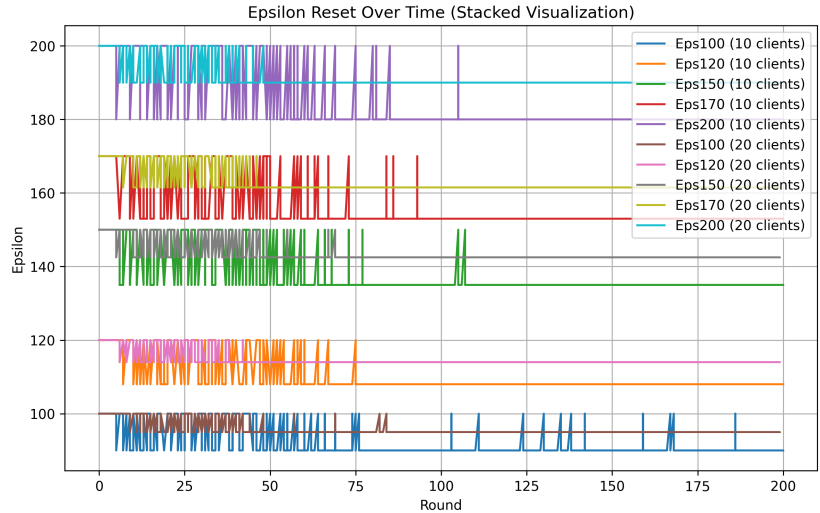

**Figure 4.6:** Effect of number of selected clients on updated privacy budget

Second, when 20 clients are selected per round, the $\epsilon$ values evolve more smoothly and steadily. This occurs because each client's individual update has a smaller impact on the overall aggregation, reducing fluctuations and making the adjustment process more stable. In contrast, with only 10 clients, the $\epsilon$ values exhibit more abrupt and frequent changes. This is due to greater variability between rounds, prompting the adaptive mechanism to adjust the noise level more often.

### 4.1.6 Effect of the Adaptive Privacy Budget Approach on Total Privacy Budget Consumption

Figure 4.7 compares the cumulative epsilon values across different target privacy budgets (100–200) and client participation settings (10 vs. 20 clients), with and without the use of an adaptive privacy mechanism. We have again used the IID dataset, and all other values remain the same as described in Table 3.2. "Sum of Mean Epsilon" is the sum of all the epsilon changes across 200 rounds of training when we use adaptive clipping, whereas "Initial DP times Total Rounds" is the initial DP value times the total number of training rounds.

The experimental results show that the "Sum of Mean Epsilon" always lies below the lines of "Initial DP times Total Rounds," indicating that the model becomes more private as more noise is added. This occurs because the adaptive privacy budget only decreases over training rounds, ensuring that the total fixed privacy budget sum remains higher than the adaptive privacy budget sum.

We observe that when the epsilon is smaller, the difference between the "Sum of Mean Epsilon" and "Initial DP times Total Rounds" is less pronounced compared to when the epsilon is larger. This is because higher epsilon values experience more significant changes in the updated value than smaller ones.

Additionally, increasing the number of selected clients reduces the impact on the difference between "Sum of Mean Epsilon" and "Initial DP times Total Rounds," as seen in Figure 4.6.

With more clients, the epsilon update jumps are smaller compared to scenarios with fewer selected clients. These findings demonstrate that adaptive privacy tuning offers substantial privacy savings while maintaining strong utility performance, aligning with the objectives outlined in cosAFed [Wan+24].

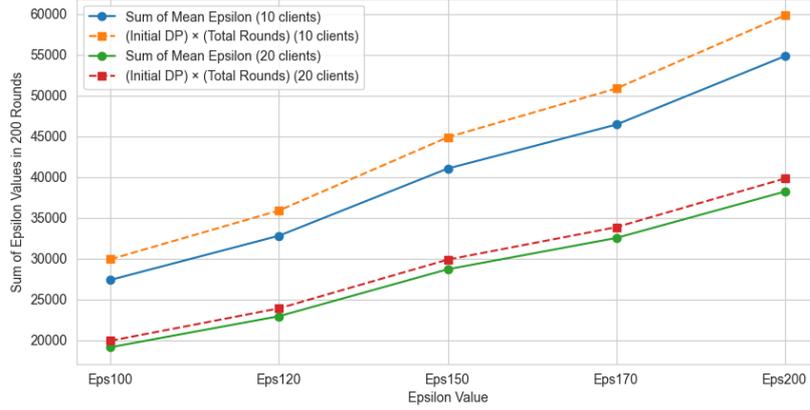

**Figure 4.7:** Different between epsilon values with and without adaptive privacy budget

## 4.2 Evaluation of APB-Gauss in Comparison with APB-Lap

In this section, we compare the Gaussian and the Laplace mechanisms with the assumption that gradients are bounded to fixed values of 1 and 0.5. The rationale behind testing with a fixed sensitivity value of 0.5 is that noise generation differs between the two mechanisms:

- In the Laplace mechanism, noise scales proportionally with sensitivity ($\Delta f$).
- In the Gaussian mechanism, noise scales quadratically $(\Delta f)^2$.

Since squaring 1 yield the same value, we instead test with a smaller sensitivity (($\Delta f$) = 0.5) to better observe the impact of noise scaling. This approach also aligns with real-world scenarios where model sensitivity tends to decrease as convergence progresses. The initial parameter values for these experiments are set according to the configurations specified in Table 3.1 and Table 3.2.

### 4.2.1 Effect of Privacy Budget on Model Accuracy Using APB-Gauss and APB-Lap

We conduct experiments with IID and Non-IID datasets with epsilon values 100, 150, and 200 and compare their mean accuracy results with Gaussian and Laplace Mechanism. In this experiment, the number of selected clients per round is 20.

As shown in Figure 4.8 and in Figure 4.9, the Laplace mechanism achieves higher mean accuracy than the Gaussian mechanism in each different $\epsilon$ value. Gaussian-IID 200 achieves lower accuracy than Laplace-IID 100, which means that increasing privacy doesn't come with accuracy cost when assuming that gradients are bound to the fixed values. It shows that in each round, the Gaussian mechanism adds more noise compared to the Laplace mechanism in each round, resulting in lower accuracy.

It is also noticeable that the Non-IID dataset gets slightly higher accuracy than the IID dataset in both mechanisms. The reason behind this is described in Section 4.1.

Figure 4.10 shows the updated privacy budget value in each round when the IID dataset is used. The selected clients in each round are 20. We compare the Laplace and Gaussian mechanisms with different privacy budget values.

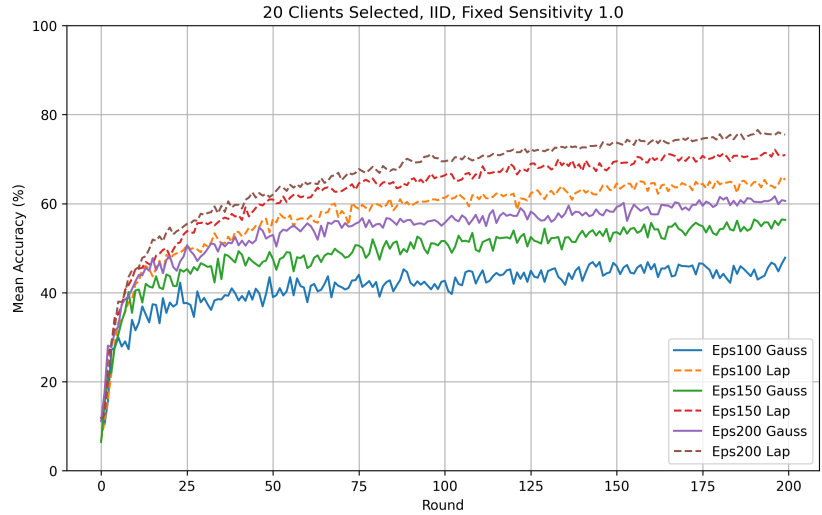

**Figure 4.8:** Mean validation accuracy for IID data using Laplace and Gaussian mechanism

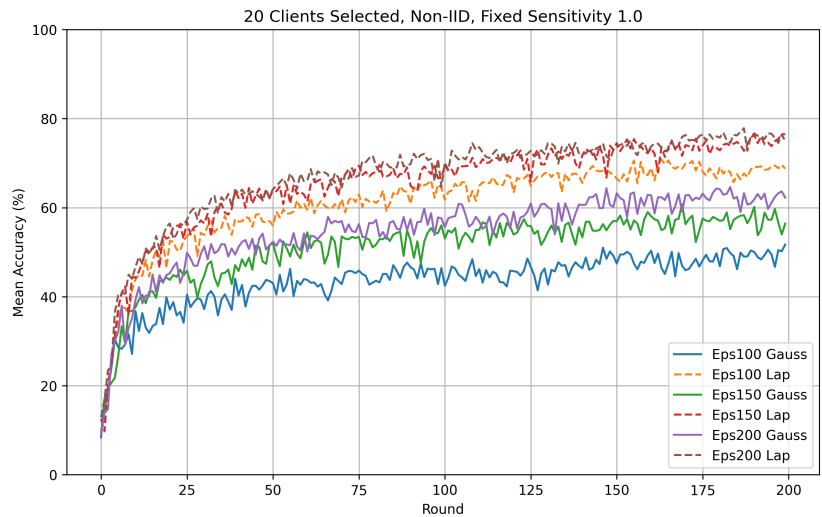

**Figure 4.9:** Mean validation accuracy for Non-IID data using Laplace and Gaussian mechanism

It shows that changing the Mechanism has similar results in the updated privacy budget in each round. This is because the updated privacy budget depends on the accuracy and loss of the global model of previous rounds, but it doesn't depend on the direct mechanism. As we are not changing the adaptive privacy budget algorithm, it doesn't have too much of an impact on the updated privacy budget.

### 4.2.2 Effect of Different Sensitivity Values on Model Accuracy Using APB-Gauss and APB-Lap

In these experiments, we examine how reducing the gradient bound from 1.0 to 0.5 affects model accuracy. We use epsilon 150 on an IID dataset, with 10 clients selected in each round. We chose 10 clients specifically to observe how adjusting the fixed bound influences accuracy, given that our earlier findings showed that increased client participation tends to improve validation performance. Finally, we compare both accuracy and loss outcomes for the Laplace and Gaussian mechanisms.

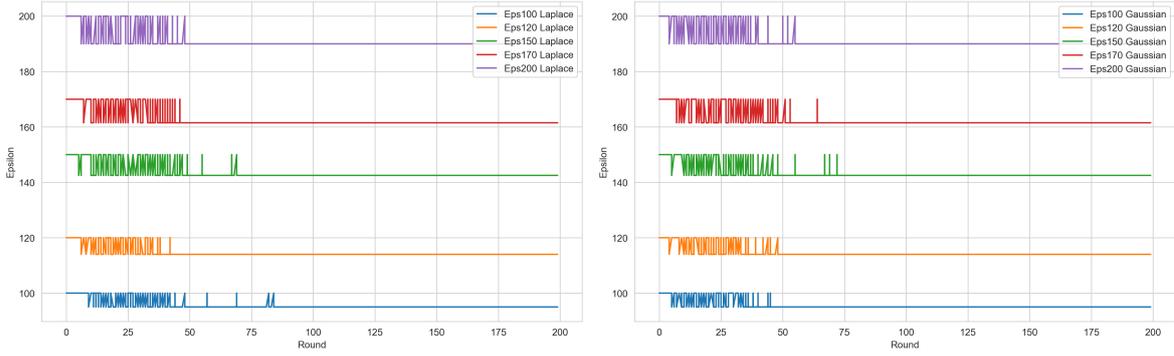

**Figure 4.10:** Updated privacy budget across rounds for varying mechanisms

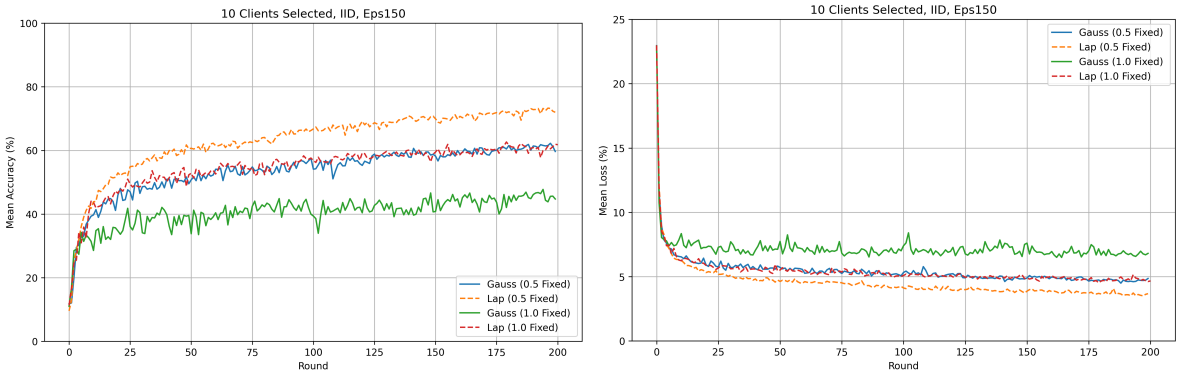

**Figure 4.11:** Mean validation accuracy and loss with varying fixed sensitivities

As shown in Figure 4.11, decreasing the fixed sensitivity values boosts the accuracy for both the Laplace and Gaussian mechanisms. When using the Gaussian mechanism, the accuracy at round 200 jumps from 45% to 60%. Meanwhile, the Laplace mechanism's mean validation accuracy increases from 62% to 72%. This improvement occurs because lowering the fixed sensitivity value reduces the noise scale, leading to less noise than with a higher fixed sensitivity value.

Notably, the Gaussian mechanism shows a 15% jump in accuracy, compared to a 10% increase for the Laplace mechanism. This happens because, in the Laplace mechanism, the noise scale is directly proportional to sensitivity, while in the Gaussian mechanism, it is proportional to the sensitivity square. When the fixed sensitivity value is below 1.0, squaring it produces an even smaller value, thereby lowering the noise scale and adding less noise when using the Gaussian mechanism.

From Figure 4.11, we can see that the Gaussian mechanism with Eps150 experiences more fluctuation than the Laplace Eps150 approach. This happens because the Gaussian mechanism struggles to stabilize in the early stages, and some clients' updates differ significantly from others, leading to these larger swings. It's also worth noting that the Laplace mechanism still achieves lower noise levels than the Gaussian mechanism when the fixed sensitivity value is reduced.

Figure 4.12 shows the final-round accuracies under various fixed sensitivity values. The Laplace mechanism demonstrates higher accuracy than the Gaussian mechanism. Additionally, reducing the fixed sensitivity below 1.0 improves accuracy for both mechanisms, as explained in the previous discussion.

Theoretically, increasing the fixed sensitivity above 1.0 would lead to lower accuracy for both Gaussian and Laplace mechanisms compared to a sensitivity of 1.0. This outcome arises

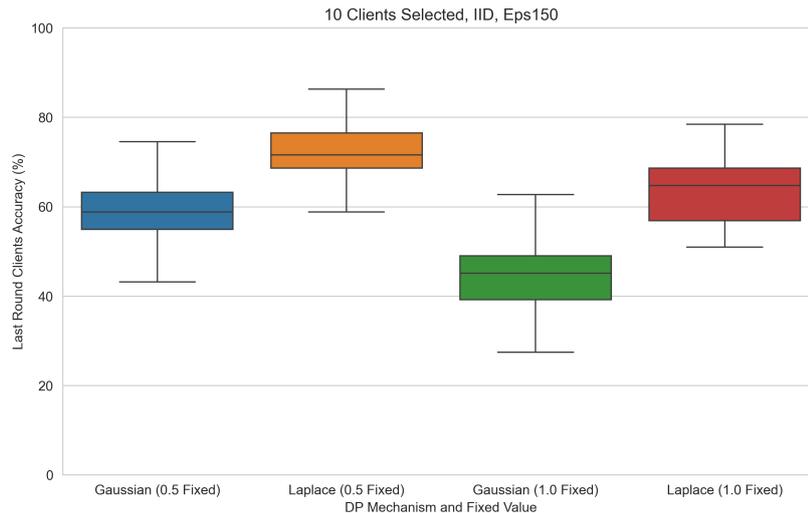

**Figure 4.12:** Final round accuracy with different fixed sensitivity

from the proportional relationship between the sensitivity value and the noise scale.

## 4.3 Evaluation of APB-GAClip in Comparison with APB-Lap and APB-Gauss

In this experiment, we compare the performance of adaptive clipping with the Gaussian mechanism against fixed clipping with the Gaussian and Laplace mechanisms. For adaptive clipping, we set the initial threshold value to 5. This choice is made because, in adaptive clipping, the threshold increases over rounds, and starting with a lower value allows for a gradual adjustment.

### 4.3.1 Effect of Data Distribution and Number of Selected Clients on Model Accuracy with APB-GAClip

We conducted experiments with privacy budgets of 100, 150, and 200 on both IID and non-IID datasets. Figure 4.13 displays results from selecting 10 clients per round on the left and 20 clients per round on the right.

With the APB-GAClip algorithm, the IID dataset consistently attains a slightly higher mean validation accuracy than the non-IID dataset. Meanwhile, Figure 4.3 indicates that without noise, the non-IID dataset achieves higher mean accuracy than the IID counterpart. This likely arises because noise has a greater destabilizing effect on the imbalanced non-IID data, causing greater fluctuations than in the IID case.

Comparing the scenarios of 10 selected clients versus 20, overall accuracy is higher with 20 clients, which suggests that using more clients can speed model learning. However, both setups exhibit pronounced fluctuation in accuracy—an indication that the model updates vary substantially from client to client even when more clients participate. Additionally, comparing the IID and non-IID curves shows elevated fluctuation in both cases, likely because APB-GAClip converges less smoothly at early rounds. Testing more rounds with a larger set of clients should be explored to optimize this clipping approach further.

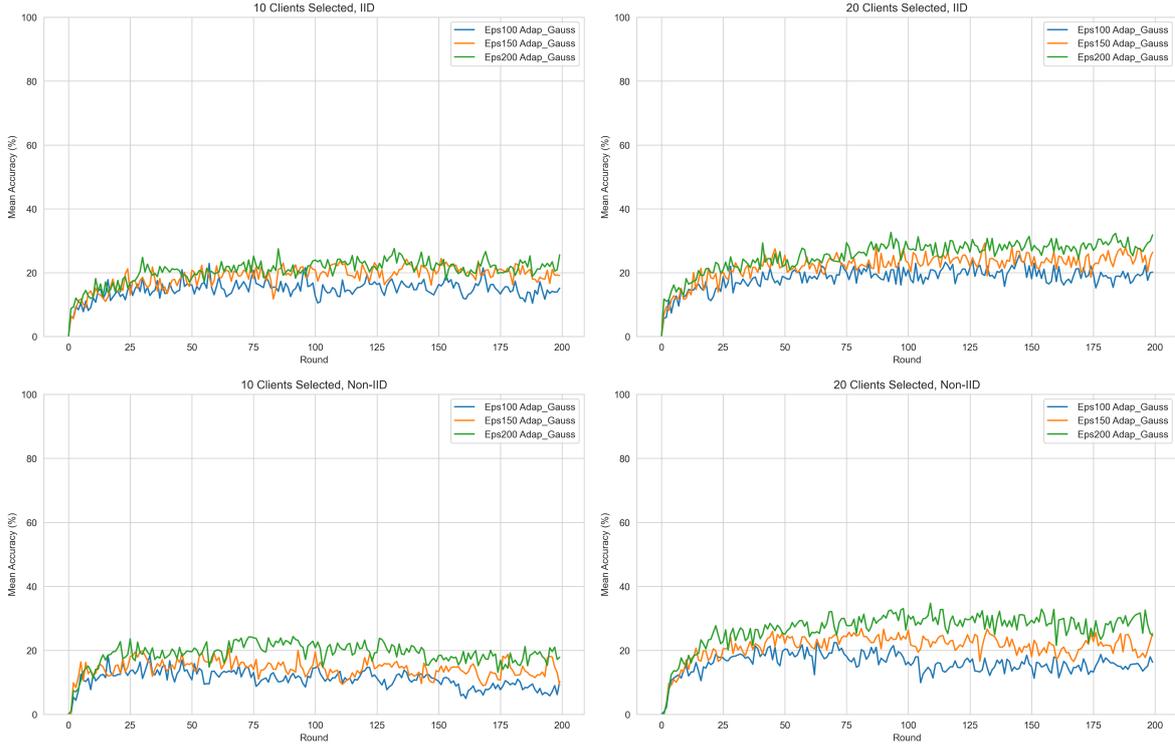

**Figure 4.13:** Effect of data distribution on model accuracy with initial threshold 5

### 4.3.2 Effect of Different Algorithms on Model Accuracy

As shown in Figure 4.14, the Laplace mechanism consistently outperforms the other two differential privacy methods, Gaussian and Adap-Gaussian in terms of final model accuracy at convergence, across all tested privacy budgets. Regardless of the specific epsilon value, Laplace noise yields the highest accuracy, suggesting it strikes a more effective balance between privacy protection and utility preservation in this federated learning setting. The Gaussian mechanism generally performs moderately, typically landing between Laplace and Adap-Gaussian. While it does converge to a lower final accuracy compared to Laplace, it still surpasses Adap-Gaussian in most scenarios. One reason for this performance gap could be that Gaussian noise, though effective, introduces a slightly higher level of randomness that hampers precise model updates, particularly under tighter privacy constraints.

Adap-Gaussian, which involves an adaptive noise injection strategy based on certain sensitivity parameters, consistently produces the lowest final accuracy. This suggests that the adaptive mechanism may be overly cautious or aggressive in estimating sensitivity, leading to higher noise levels than necessary. As a result, the learning process becomes more noisy and unstable, causing the model to underperform relative to its non-adaptive counterparts.

Moving from lower epsilon 100 to higher epsilon 200 generally increases the final accuracy for all three noise methods. A large epsilon means less overall noise, so the models can fit the data more closely and reach higher accuracy. Because the data are non-IID across the 20 clients, more noise is usually needed, and thus, each noise mechanism imposes a distinct accuracy penalty. The difference between Laplace and Gaussian is especially visible here, showing that Laplace with the same epsilon may introduce effectively less damaging noise in this setup.

Overall, Laplace noise injection achieves the best accuracy at every privacy budget tested, followed by standard Gaussian, and then Adap-Gaussian is substantially lower. Moreover, increasing epsilon reduces noise enough that all methods converge to higher accuracy.

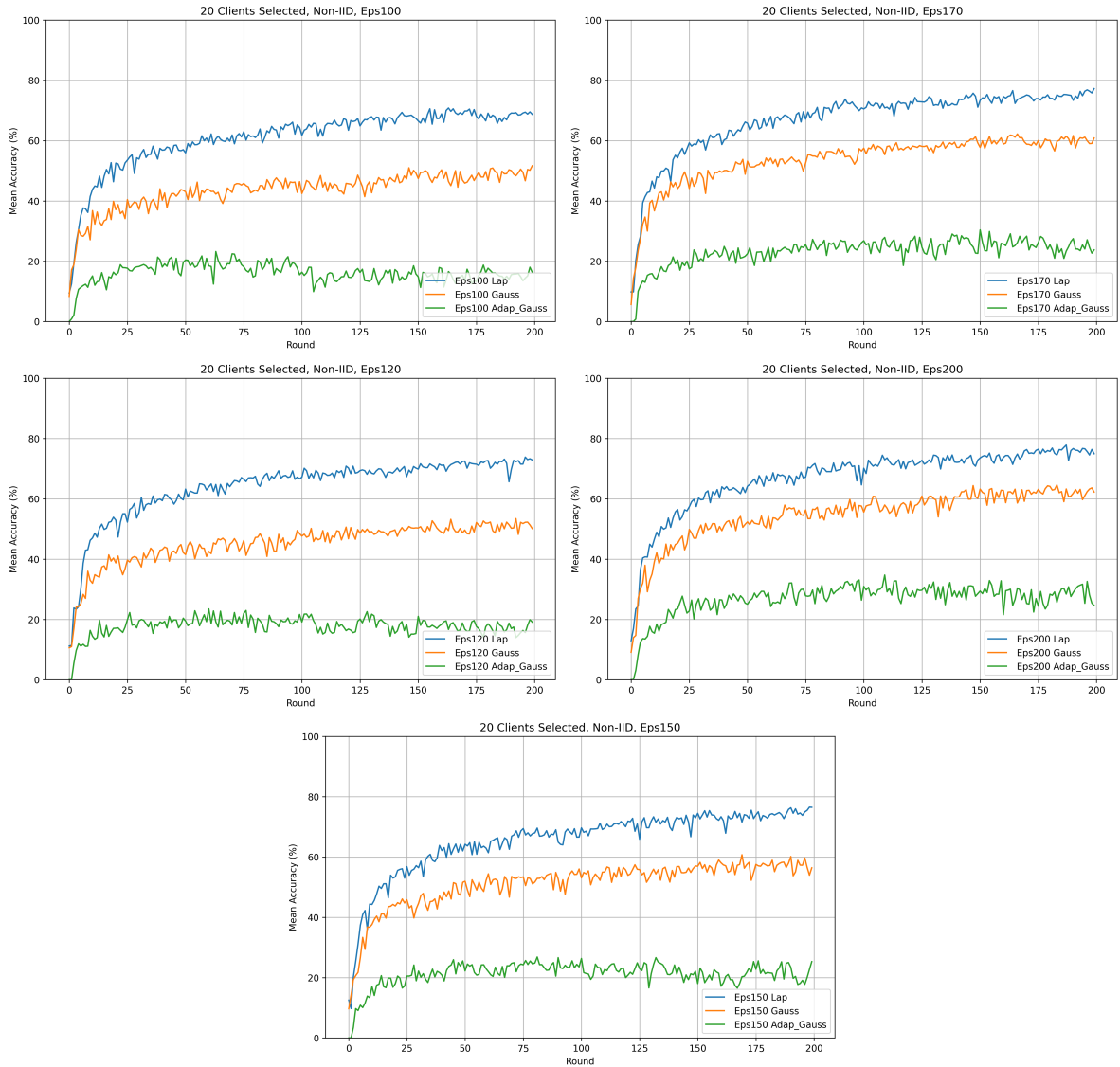

**Figure 4.14:** Effect of different algorithms on model accuracy

### 4.3.3 Effect of Number of Selected Clients on Updated Privacy Budget Using APB-Lap and APB-GAClip

From the below Figure 4.15, we experimented to see how the Adap-Gaussian mechanism (left side) versus the Laplace mechanism (right side) adjusts the effective privacy budget round by round under non-IID dataset, comparing 10 clients selected and 20 clients selected.

Each epsilon value's line hovers near its initial epsilon value on the vertical axis, indicating the noise mechanism is trying to stay around that target privacy level. The Adap-Gaussian shows larger swings in epsilon from round to round. This is because the algorithm dynamically tunes the noise scale in each round based on accuracy, loss, etc. Whereas on the right side figure shows that it still has some smaller shifts, the Laplace-based lines are more stable. The clipping and the noise addition produce fewer abrupt epsilon jumps each round.

With 10 clients selected, the adaptive lines tend to oscillate more dramatically. Smaller client subsets per round can introduce larger differences in gradient norms, causing bigger swings in the adaptive epsilon. When 20 clients are selected, the adaptive lines still vary, but often less wildly. Having more client updates each round can average out extremes in the data or gradients, resulting in somewhat steadier adjustments.

Overall, these figures show that Adap-Gaussian attempts to modulate the effective privacy budget more responsively each round, but at the cost of noticeable epsilon volatility—especially with fewer selected clients. In contrast, Laplace-based noise stays nearer to its nominal budget value, indicating more uniform clipping/noise behavior from round to round.

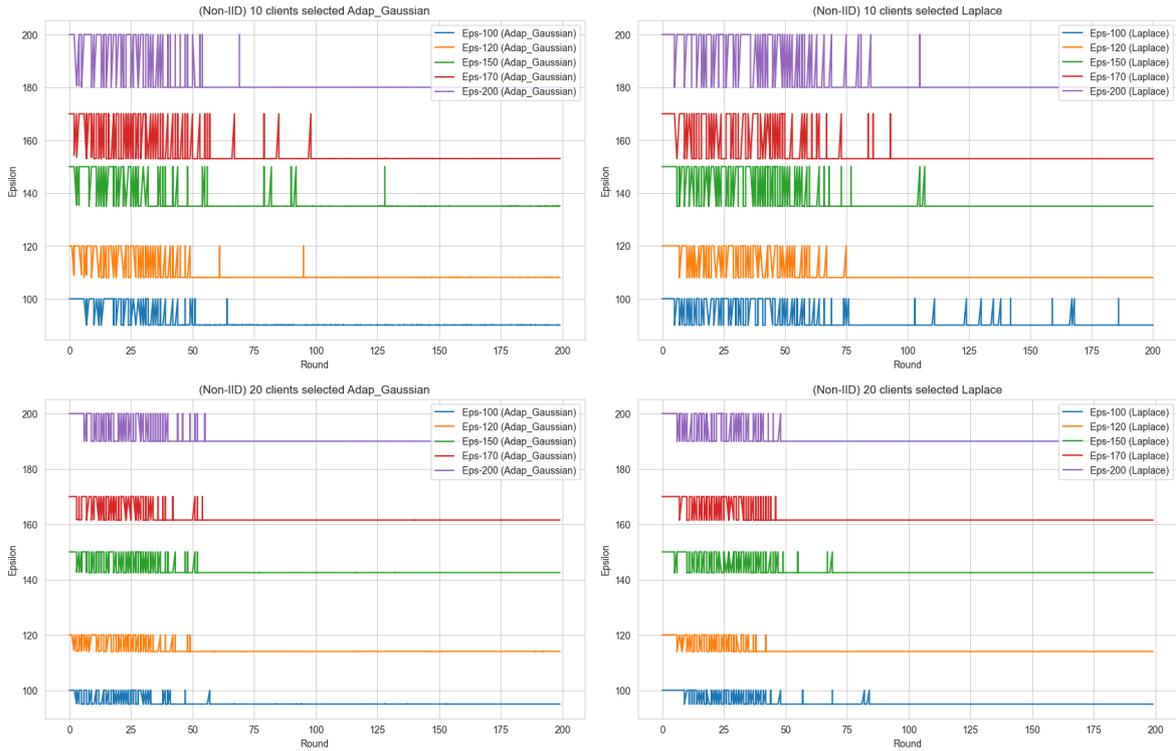

**Figure 4.15:** Effect of number of selected clients on updated privacy budget using APB-Lap and APB-GAClip

# Chapter 5

# Discussion

## 5.1 Summary of Key Findings

This study evaluate the three differential privacy (DP) mechanisms (APB-Lap, APB-Gauss, and APB-GAClip) in federated learning (FL) using the CIFAR-10 dataset under different data distribution setting. Key results includes:

1. APB-Lap achieves higher accuracy compared to Gaussian-based methods.

2. With fixed sensitivity ($\Delta f = 0.5$) achieves better accuracy compare to ($\Delta f = 1.0$) with APB-Lap and APB-Gauss.

3. Non-IID data distribution (Dirichlet, $\alpha = 0.5$ for label and quantity distribution) achieves higher baseline accuracy (85%) than IID (78%). But when the DP noise is added, IID data achieves slightly better accuracy than Non-IID data.

4. Increasing the number of clients selections improved the model accuracy by reducing the noise.

5. Adaptive privacy budget has more influence from the global model metrics rather than noise type.

6. APB-GAClip underperformed due to instability from the adaptive clipping threshold and limited training duration.

Gaussian mechanism are often preferred for high-dimensional data due to smoother noise distribution, even though APB-Lap achieves stronger performance. This may be because of three reasons. First, we assume the gradients are bounded ($\Delta f = 1.0$), which simplify the noise scaling as added noise is proportional to $\Delta f$. In Gaussian noise, added noise is proportional to $(\Delta f)^2$, which leads to generate more noise even at moderate $\Delta f$ values. Secondly, Laplace's sharper noise decay is exponential, where are Gaussian noise decay is quadratic. With 200 rounds, Laplace convergence faster and Gaussian mechanism requires longer training to stabilize. Thirdly, we use CIFAR-10 dataset, which has moderate dimensionality, but gaussian noise is achieves better results in high-dimensional spaces. This finding suggests that Laplace mechanism are pragmatic for short-term FL tasks with bounded gradients, while Gaussian methods may require extended training.

Non-IID data usually gives better baseline performance. Because clients focusing on fewer classes helped improve the global model. However, when DP noise is added, non-IID data becomes more unstable. The uneven updates from clients make the model more sensitive to noise. This shows that we may need to adjust the noise carefully depending on the data, stronger privacy could come at the cost of losing the benefits of diverse data.

We used an adaptive privacy budget that changed based on model performance (like accuracy or loss). This helped save the total privacy cost by only adding noise when needed. Importantly, this method worked the same no matter what kind of DP noise (Laplace or Gaussian) was used, since it relied on overall model performance, not specific noise settings. This makes adaptive systems like cosAFed [Wan+24] flexible and useful in many DP scenarios.

The lower accuracy seen with APB-GAClip shows that adaptive clipping can struggle when training for a short time. Starting with a high threshold value 5.0 might have added too much noise early on. Also, the method used to adjust the threshold (EMA) didn't settle properly within 200 rounds.

## 5.2  Future Research Directions

In our current work, adaptive clipping was applied only with the Gaussian mechanism. In the future, both fixed and adaptive clipping strategies can be explored for the Laplace mechanism as well. Additionally, the Gaussian mechanism can be evaluated with a fixed clipping value. Although we performed 200 training rounds, conducting more rounds could yield better insights, especially when using the Gaussian mechanism. Rather than relying solely on CIFAR-10, future work can incorporate higher-dimensional datasets for more comprehensive evaluation. Our experiments were limited to the FedProx and FedAvg aggregation strategies. However, the *SelecEval* framework can be extended to support other aggregation methods with differential privacy, such as FedNova, FedDisco, and FedAvgM. Currently, we use an adaptive privacy budget based on the global model's accuracy from previous rounds. This approach can be further extended to adapt privacy budgets based on individual clients' data sensitivity, allowing more personalized privacy preservation.

# Appendix A

# Appendix

## A.1 Different Terms in Differential Privacy

- **Fixed Threshold**: The upper limit of the gradient norm. For example, if the fixed threshold is 1.0 and the gradient norm is 2.5, it will be scaled down to 1.0 before adding differential privacy (DP) noise.

- **Fixed Clipping**: Another term for Fixed Threshold, commonly used in gradient norm clipping.

- **Adaptive Clipping**: Dynamically adjusts the clipping threshold value based on the gradient distribution over time.

- **Clipping Threshold Setting**: "We set the clipping threshold to 1.0 to prevent large gradients from dominating the updates."

- **Adaptive Threshold Adjustment**: An adaptive clipping threshold adjusts over time based on the gradient distribution.

- **Gradient Norm Clipping**: If the gradient norm exceeds a predefined threshold, it is clipped to bring it within the allowable limit.

- **Pre-Noise Clipping**: Before adding noise to ensure differential privacy, the gradient norm is clipped to control the sensitivity.

- **lr_threshold**: This parameter is used to update the clipping threshold on the server side. The threshold can be defined in various ways, such as based on gradient norms, model sensitivity, or a fixed clipping value.

- **lr_sensitivity**: During noise addition in model updates, this parameter is used to represent the model sensitivity. In adaptive clipping, the upper bound of sensitivity is adjusted using the lr_threshold. To prevent excessive updates to the threshold, lr_sensitivity is applied when computing the noise scale.